\documentclass[runningheads,a4paper]{llncs}

\usepackage{amsmath}
\usepackage{amssymb}
\usepackage{soulutf8}
\usepackage{xcolor}
\usepackage[dvipdfmx]{graphicx}
\usepackage{bmpsize}
\usepackage{amsmath}
\usepackage[ruled]{algorithm2e}
\usepackage{bm}
\usepackage{xcolor}
\usepackage{transparent} 
\usepackage{fixltx2e}
\usepackage{epstopdf}

\usepackage[normalem]{ulem}
\usepackage[export]{adjustbox}
\DeclareGraphicsExtensions{.eps,.pdf,.png,.jpg}

\newif\ifanonymise
\anonymisefalse

\newcommand{\doubleblind}[1]{
\ifanonymise
{\noindent[ identifying information removed ]}
\else
{#1}
\fi
}
\let\normalauthor\author
\renewcommand{\author}[1]{\normalauthor{\doubleblind{#1}}}
\let\normalinstitute\institute
\renewcommand{\institute}[1]{\normalinstitute{\doubleblind{#1}}}

\newif\ifabstractonly
\abstractonlyfalse

\newif\ifshowwip
\showwiptrue
\showwipfalse

\newcommand{\tocite}[1]{\textcolor{red}{[x]}}

\newcommand{\pagelimit}[1]{
\ifshowwip
\ifnum\thepage>#1
	\noindent\textcolor{red}{\bf STOP! TOO MUCH!}
\fi
\fi

}

\ifshowwip
\usepackage[outer]{showlabels,rotating}

\fi

\newcommand{\SatO}[0]{\text{SO$_2$}}
\newcommand{\HbO}[0]{\text{HbO$_2$}}
\newcommand{\Hb}[0]{\text{Hb}}
\newcommand{\THb}[0]{\text{THb}}

\makeatletter
\newcommand{\removelatexerror}{\let\@latex@error\@gobble}
\makeatother

\usepackage{url}
\newcommand{\keywords}[1]{\par\addvspace\baselineskip
\noindent\keywordname\enspace\ignorespaces#1}

\soulregister\cite7
\soulregister\ref7
\definecolor{lightBlue}{rgb}{.7,.90,1}
\definecolor{paleYellow}{rgb}{1,.97.,.4}

\makeatletter
\ifdefined\disablehighlights

\else

\fi
\makeatother

\makeatletter
\ifdefined\disablehighlights

\else

\fi
\makeatother

\makeatletter
\newcommand{\stfix}[1]{\relax}
\makeatother

\begin{document}

\mainmatter  

\title{Fast Estimation of Haemoglobin Concentration in Tissue Via Wavelet Decomposition}

\titlerunning{Fast Haemoglobin Concentration Estimation by Haar Decomposition}

%
%
\author{Geoffrey~Jones \and Neil~T~Clancy \and Xiaofei~Du \and Maria~Robu \and Simon~Arridge \and Daniel~S~Elson \and Danail~Stoyanov}

\authorrunning{Jones et al}

\institute{}

%
%

\maketitle

\begin{abstract}

Tissue oxygenation and perfusion can be an indicator for organ viability during minimally invasive surgery, for example allowing real-time assessment of tissue perfusion and oxygen saturation.
Multispectral imaging is an optical modality that can inspect tissue perfusion in wide field images without contact.
In this paper, we present a novel, fast method for using RGB images for MSI, which while limiting the spectral resolution of the modality allows normal laparoscopic systems to be used.
We exploit the discrete Haar decomposition to separate individual video frames into low pass and directional coefficients and we utilise a different multispectral estimation technique on each.
The increase in speed is achieved by using fast Tikhonov regularisation on the directional coefficients and more accurate Bayesian estimation on the low pass component.
The pipeline is implemented using a graphics processing unit (GPU) architecture and achieves a frame rate of approximately 15Hz.
We validate the method on animal models and on human data captured using a da Vinci stereo laparoscope.

\end{abstract}

\section{Introduction}

Haemoglobin concentrations in tissue are important measurements that provide functional information and could also be used for structural characterisation of tissue types.
In minimally invasive surgery (MIS), the use of haemoglobin concentration imaging on tissue surfaces using multispectral imaging (MSI), is promising as an non-ionising optical imaging modality that can monitor organ viability in transplant procedures \cite{Clancy2016Uterine}, \cite{Clancy2015:BowelOx} or be used to detect abnormal tissue.
A major advantage of the MSI modality is that it is able to obtain wide field of view measurements without contact thus allowing for monitoring large areas that cannot be observed through other sensing means like oxi-meter probes which can only give spot measurements.

Techniques for MSI compatible with MIS are typically limited for real-time monitoring by either their capture rate \cite{Clancy12}, data processing speed \cite{Jones17} limiting their use for imaging dynamic systems.
Methods have been developed to estimate tissue haemoglobin concentrations for monitoring using fast filter wheels of minimal sets of filters that show significant promise \cite{Wirkert2014}.
Yet in MIS the surgical environment makes the use specialised hardware complex because it requires additional regulatory and sterilisation considerations to be accounted.
As such computational methods requiring minimal hardware modification and using existing hardware are highly attractive.
Computational techniques that utilise the laparoscopic RGB video feed for tissue monitoring at speeds greater than 15Hz, can require inflexible calibration \cite{Nishidate2013}, or has difficulty dealing with scene motion \cite{Guazzi2015}, or requires making a trade off in estimation accuracy of saturation (\SatO) against total haemoglobin (\THb) \cite{Jones2016Tik}.
Therefore high-speed techniques for this purpose are still a major challenge. 

Wavelet decompositions have been widely used in image processing to transform data it into domain spaces where certain operations can be applied more efficiently or components of the signal with little entropy can be omitted \cite{Daubechies1992}.
In this paper, we adopt the wavelet approach and propose a framework for processing frames of laparoscopic video by Haar decomposition, which allows us to use separate algorithms to process the various components of the compressed data representation.
Because the Haar wavelet produces large numbers of zeros in the transformed data set, information is concentrated in relatively few coefficients which we can process effectively.
The resulting algorithm means we can approach RGB processing to arrive at a surrogate MSI signal though a dual optimisation approach which seems effective in fairly smooth signals, such as laparoscopic video data.
We present preliminary results on data captured using hardware MSI for comparison on animal models as well as results on human tissue imaging acquired within a laboratory environment.

\section{Method}

Underpinning the technique proposed in this paper is the estimation of a signal surrogate to MSI but obtained from RGB video frames.
This relies on knowledge of the tissue's transmitted attenuation characteristics \cite{Bosschaart2013}, backscattered attenuation coefficients are derived by Monte Carlo (MC) simulation \cite{MMCFang2010}, to combine attenuation due to absorption and scattering \cite{Jones17}, since camera and light source are on the same side of the tissue during laparoscopic imaging.
The fitting to surrogate MSI data can be performed to satisfy the Beer-Lambert relation \cite{Clancy12}.

Once an RGB image is captured, the 2D discrete Haar transform can be used to decompose the image into four components, three directional approximate derivatives and a residual low pass coefficient.
We exploit this fact for efficient computation since it enables us to only perform expensive estimation on less of the data ($1/2^{1+n}$ where $n$ is the level of decomposition).
The matrix $H$ can be expressed for an image window $I = \{I_{x,y},I_{x,y+1},I_{x+1,y},I_{x+1,y+1}\} $ as:

\begin{equation}
  \text{Haar}(I) = I*H
\end{equation}
given that the 2d Haar matrix is formed as:
\begin{equation}
  H = 0.5\cdot
  \begin{bmatrix}
    1 &  1 &  1 &  1\\
    1 &  1 & -1 & -1\\
    1 & -1 & -1 &  1\\
    1 & -1 &  1 & -1
 \end{bmatrix}
\end{equation}

As this is a just a linear transformation, if we group each pixel data in windows of four elements we can include it in the construction of RGB data $I_{RGB}$ from the multispectral signal $I_\lambda$:

\begin{align}
C\cdot I_\lambda \cdot H &= I_{RGB} \cdot H\\
\intertext{for known camera spectral sensitivity matrix $C$, the least squares solution being:}
\label{e:hwavesolve}
I_\lambda &= ((C^\text{T} C)^{-1} C^\text{T} \cdot I_{RGB} \cdot H) \cdot H\\
\intertext{since $H = H^*$.
Similarly for the Tikhonov regularise solution as used by \cite{Jones2016Tik}, we can apply this method to the Haar transformed data as}
I_\lambda \cdot H &= (C^\text{T} C + \Gamma)^{-1} C^\text{T} \cdot I_{RGB} \cdot H
\end{align}
where $\Gamma$ is the identity matrix multiplied by a small constant $\gamma$.
Here we have used the same windowed grouping:
\begin{align}
I_{RGB} &= \{I_{RGB,x,y}, I_{RGB,x,y+1},I_{RGB,x+1,y},I_{RGB,x+1,y+1},\}\\
\intertext{corresponding to the multispectral window:}
I_\lambda &= \{I_{\lambda,x,y},I_{\lambda,x,y+1},I_{\lambda,x+1,y}, I_{\lambda,x+1,y+1},\}
\end{align}

While this demonstrates how Tikhonov based estimation can be used on any of the Haar coefficients, to get better accuracy it is desirable to only use this on the sparse directional components.
On the low-pass coefficients we then utilise the Bayesian method of \cite{Jones17}, which they show is more accurate then Tikhonov estimation, enabling estimation of oxygen saturation.
It is possible to do so because the low pass coefficients are analogous to an over exposed image taken at a lower resolution.

Formally we can show that the method holds by substituting in the maximisation step of \cite{Jones17} with low pass transformed data.
Using $H_{LP} = 0.5*[1,1,1,1]$ to represent the low pass component of the discrete Haar transform, starting from the concentration estimation $\hat{x}$ from low pass data:
\begin{align}\label{e:concentrationE}
\xi x &= -log(I_\lambda H_{LP})\\
\intertext{where $\xi$ are the backscatter attenuation coefficients for oxy and de-oxy haemoglobin (\HbO\ and \Hb) as well as a constant term.
The constant term is included to account for global illumination changes that occur equally across wavelength, such as due to distance from the camera.
We solve for $x$ using least squares fitting,}
\hat{x} &= (- \xi'\xi)^{-1}\xi'\log(I_\lambda H_{LP})
\end{align}
These concentration estimates are then used to generate the expected value for the multispectral data $E[I_\lambda]$ used to regularise then subsequent fitting iteration, this is defined as,
\begin{align}
E[I_\lambda] &= e^{-\xi x}\\
\intertext{substituting the concentration estimate from \eqref{e:concentrationE} and cancelling terms the expected value will be the multispectral data transformed by $H_{LP}$}
E[I_\lambda] &= I_\lambda H_{LP}
\end{align}
This can only apply to the low pass coefficients because they are all positive and so all terms remain real throughout.

The expectation step of \cite{Jones17} is linear and so compatible with working on Haar transformed data.
One notable change that we make to the expectation step is to change the prior on the expected spectrum from a value prior to a shape prior.
We do this by computing the second derivative of $E[I_\lambda]$ and using that to regularise the second derivative of the estimated $\hat{I}_\lambda$, this better regularises the estimation.

\section{Experiments and Results}

We conducted two experiments to validate the proposed algorithm.
The first uses RGB views generated from {\em in vivo} animal experiment MSI data sets, which allow evaluation comparing the method to a hardware implemented gold standard.
Secondly we utilise our method to process video from a patch of tissue at the base of the tongue, where we show that because we are able resolve at a high frame rate we can detect the pulse rate by tracking a patch of tissue over time.

We observed good stability of our method both on the generated RGB data and on the in-vivo acquired RGB data.
This is conformant to expectations as our method independently combines two methods which either have a closed form solution \cite{Jones2016Tik} or are empirically observed to be stable under noisy conditions \cite{Jones17}.

\begin{figure}[t]
\begin{minipage}{\linewidth}\centering
\includegraphics[width=0.97\textwidth]{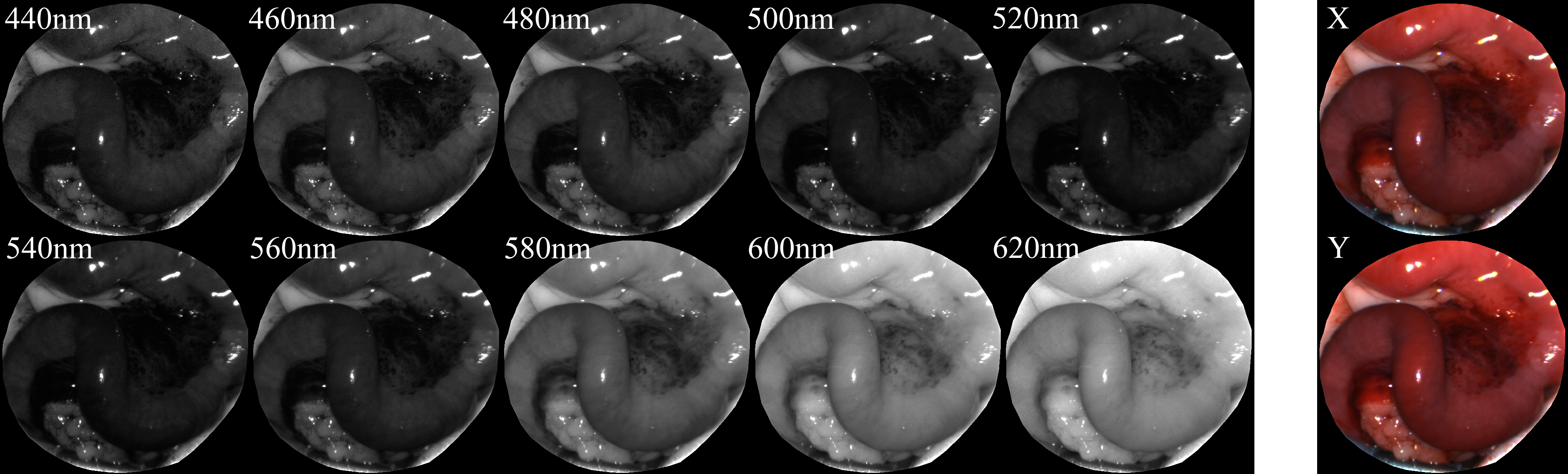}
\end{minipage}

\vspace{.5em}
\begin{minipage}{\linewidth}\centering
{\hspace{6em}{\THb\ (g/litre)} \hfill{\SatO\ (\%)}\hspace{7em}}

\includegraphics[width=\textwidth,trim={0 1cm 0 .6cm},clip]{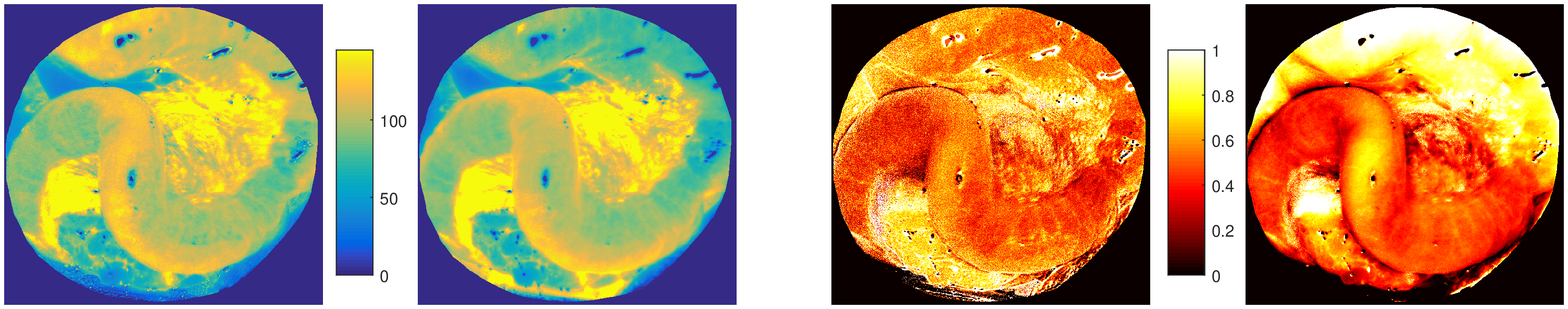}

{\hfill
{reference \hspace{4.6em} proposed}
\hspace{5em}
{reference \hspace{4.6em} proposed}
\hfill}
\end{minipage}
\caption{\label{f:synthetic} Top: Selected bands from a multispectral datacube and corresponding synthesised RGB views. Bottom left: total haemoglobin. Bottom right: oxygen saturation. Example estimation from our proposed method using RGB data synthesised from MSI and compared with reference estimation directly from the MSI data.}
\end{figure}

\subsection{Comparison with Hardware Multispectral Imaging Signals}

Real {\em in vivo} MSI data was used to generate synthetic RGB images using known camera spectral sensitivity curves, calibrated from the spectral response of cameras on a Da Vinci surgical robot \cite{Robu2015}.
The MSI data comes from experiments in which MSI was used to monitor and evaluate organ viability, by measuring tissue perfusion and oxygen saturation, throughout uterine transplant procedures performed on sheep and rabbits \cite{Clancy2016Uterine}.
The data was chosen as they most closely resemble the envisioned clinical use for our method.
To evaluate computation speed and accuracy we compared against \cite{Jones2016Tik} and \cite{Jones17}, however, to make fair comparisons of speed on the same architecture we also implemented both of these methods for GPU, using the CUDA programming language.
We performed two variants of our proposed method corresponding to single (w1) and multilevel (w3: three levels) Haar decomposition.
All methods' accuracy are compared against the results of estimation from directly using the MSI data \cite{Clancy12}.

\begin{table}[t]
\def\arraystretch{1.2}
\setlength\tabcolsep{1em}
\centering
\caption{\HbO\ and \Hb\ estimation accuracy on synthetic data compared to direct estimation from multispectral data \cite{Clancy12}.}
\label{t:synthetic}
\begin{tabular}{l|c|c|c}
\hline
Method & Computational & Execution time & Mean squared error  \\
       & architecture  & Hz             & g/litre             \\
\hline
Bayes \cite{Jones17}         & CPU  & 0.0671 & 25 \\ 
Tikhonov \cite{Jones2016Tik} & CPU  & 0.282  & 54 \\ 
Bayes \cite{Jones17}         & CUDA & 3.54   & 25 \\ 
Tikhonov \cite{Jones2016Tik} & CUDA & 43.5   & 54 \\ 
Proposed (w1)                & CUDA & 12.7   & 36 \\ 
Proposed (w3)                & CUDA & 14.4   & 36 \\ 
\hline
\end{tabular}
\end{table}
As seen in \figurename\ \ref{f:synthetic} the estimation from our method bears strong visual similarity to the result of direct estimation from MSI data, with two notable variations.
Firstly in our method there is much less noise in the estimation, which can be interpreted as a product of the MSI data’s band limited images each being moderately noisy.
Such noise is inherently smoothed out when synthesising RGB data from the MSI datacube as each RGB band integrates over many MSI bands due to wider sensitivity of the respective colour channels.
On the other hand, our RGB technique is less able to robustly estimate haemoglobin concentrations in areas of high illumination, this can be clearly seen in the \SatO\ estimation near the top. 
This is possibly due to the non-linear response of the camera at extreme measurements. 
In addition, the generally smoother appearance of our result means that local variations in the MSI signal are lost due to both the synthesis of the RGB image and subsequently due to the regularisation in our estimation approach.
However, overall our proposed method was closer to that of \cite{Jones17} than \cite{Jones2016Tik} as is shown in \tablename\ \ref{t:synthetic}.

In terms of computational performance, the timed results of haemoglobin estimation were made using mega-pixel sized input images.
Timings were calculated including all pre-processing steps, which for our proposed method included the initial Haar decomposition and final recomposition, this illustrates the diminishing returns for the for multileveled (w3) variant of our method compared to the single decomposition level version (w1) as tabulated in \tablename\ \ref{t:synthetic}.
Interestingly both variations have the same error performance albeit with minor computational variations in terms of speed.

\subsection{Experiments on In vivo Data From a Stereo-laparoscope}

\begin{figure}[t]\vspace{-1em}
\begin{minipage}{\linewidth}\centering
\hspace{.5em}\includegraphics[width=0.95\textwidth]{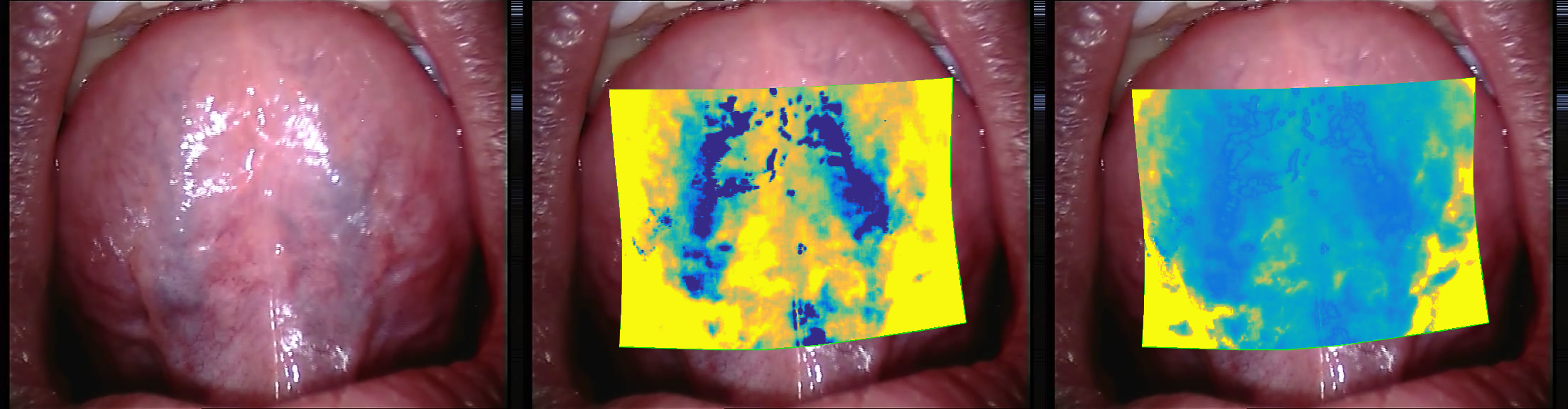}

\hspace{.5em}{\hspace{6.5em} {(a)}\hfill {(b)}\hfill {(c)}\hspace{5em} }
\end{minipage}

\begin{minipage}{\linewidth}\centering
\includegraphics[width=\textwidth]{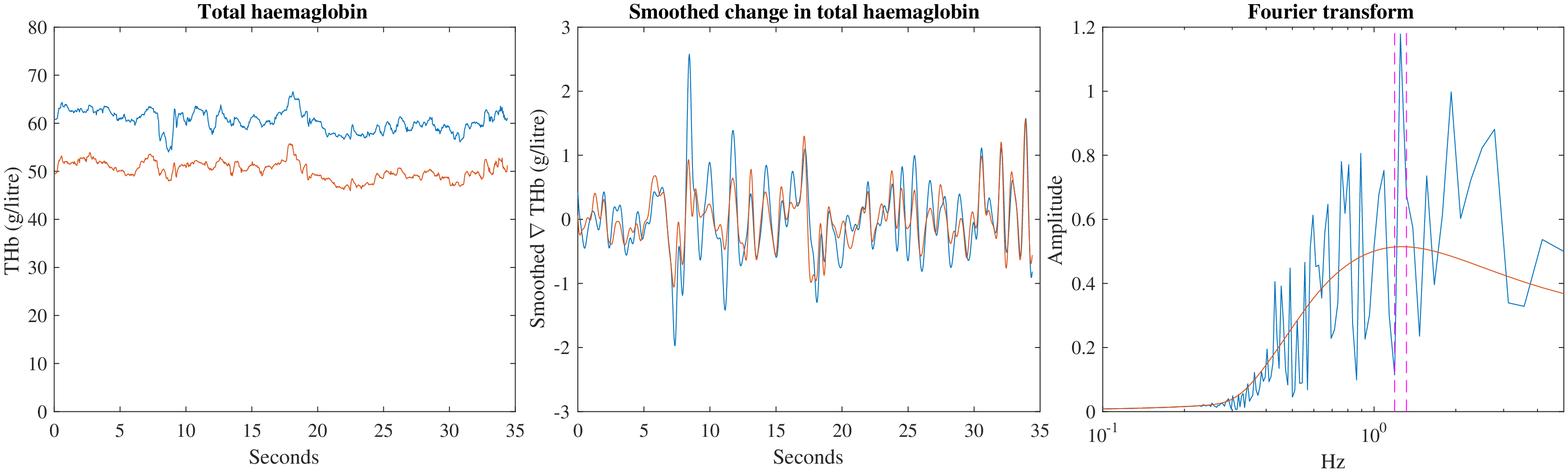}
\end{minipage}
\caption{\label{f:timegraphs}
Top (a) the original RGB laparoscopic view, (b) \SatO\ estimation overlaid on the laparoscopic view, (c) \THb estimation overlaid on the laparoscopic view.
Bottom left: the tracked \THb\ over time from the left and right camera feeds separately this is nice and constant (not bleeding from my tongue!).
Bottom centre: the derivative after smoothing the \THb\ trace for each camera view, smoothing is required due to large amounts of high frequency noise in the trace. (roaming high-lights on tissue surface, non-perfect tracking)
Bottom right: Frequency analysis of the change in \THb\ over time, fitting a polynomial curve finds the peak between between $1.2$ and $1.3$ Hz.
}\vspace{-1em}
\end{figure}
We captured video of the base of the tongue of an adult male using a Da Vinci surgical robot's stereo laparoscope as shown in Figure 2.
The laparoscopic camera was calibrated to estimate the channel sensitivity.
We tracked a patch of tissue using the method of \cite{Du2015} over time in both left and right cameras, and processed the two views separately.
Tracking and image warping was necessary in order to remove residual motion artefacts due to tongue movement allowing visualisation of the spectral and temporal variation of a selected region within a spatio-temporally registered signal.
We then compared the mean value of total haemoglobin as estimated from each view as shown in \figurename\ \ref{f:timegraphs} and observed a strong similarity in estimation over time.
While there appears to be constant offset between the estimate from either view this could be due to a miss calibration of either of the cameras' spectral response curve.
We observed a large amount of high frequency noise in the time series data, which could be due to specular highlights moving as the tissue surface deformed and moved or to imperfect tracking which is to be expected given the difficult conditions.

Processing the registered video signal after warping using a low pass filter we were able to observe a periodic signal in the derivative of the time series which was aligned in both views.
We believe this to be representative of physiological signals within the tissue due to cardiovascular activity.
By looking at the Fourier transform of the change in \THb\ estimation the power spectrum peaked between $1.2$ and $1.3$ Hz which corresponds to periodic signal of between 72 and 78 cycles per minute.
This resonated with the heart rate of the subject during the experimental acquisition.

\section{Conclusion}

We have presented a hybrid method for estimating haemoglobin concentration from RGB images.
The method improves on the computational speed, with minimal loss in accuracy.
The speed improvement is sufficient to enable use in real-time clinical applications, as it is fast enough to resolve variations in the oxy- and de-oxygenated blood flow within tissue close the exposed surface.
During surgical procedure heart rate can become elevated to over 2 Hz~\cite{bensky2000dose} as such imaging at a rate faster than 4Hz is would be required to detect this without aliasing.
The precision of the proposed method is also high enough to evaluate the oxygen saturation within tissue, this is a major improvement on~\cite{Jones2016Tik}.
Our results on {\em in vivo} measurements of haemoglobin concentrations using an existing laparoscopic set up for white light RGB imaging are promising.
Meanwhile the results on data from animal transplantation experiments illustrates the potential surrogate use to a hardware MSI approach as a translatable method for monitoring and evaluating organs intra-operatively.
Additional controlled studies are required to validate the precise nature of the recovered information and to explore the quantitative experimental differences between the proposed approach and set ups that use narrow bands of wavelengths that allow a more well constrained regression.

\section{Acknowledgements}
\vspace{-0.4em}
This work was supported by the EPSRC (EP/N013220/1, EP/N022750/1, EP/N027078/1, NS/A000027/1, EP/P012841/1), The Wellcome Trust (WT101957, 201080/Z/16/Z) and the EU-Horizon2020 project EndoVESPA (H2020-ICT-2015-688592).
\vspace{-.4em}
\bibliographystyle{splncs03}
{
\bibliography{IEEEabrv,mybib}
}

\end{document}